\documentclass[10pt,twocolumn,letterpaper]{article}

\usepackage{wacv}
\makeatletter
\@namedef{ver@everyshi.sty}{}
\makeatother
\usepackage{tikz}
\usepackage{times}
\usepackage{epsfig}
\usepackage{graphicx}
\usepackage{amsmath}
\usepackage{amssymb}
\usepackage{booktabs}
\usepackage[accsupp]{axessibility}  
\usepackage{graphicx}

\usepackage{tikz}
\usepackage{comment}
\usepackage{amsmath,amssymb} 
\usepackage{color}
\usepackage{xspace}

\usepackage[accsupp]{axessibility}  

\usepackage{times}
\usepackage{epsfig}
\usepackage{graphicx}
\usepackage{amsmath}
\usepackage{amssymb}
 \usepackage{booktabs}
\usepackage[numbers,sort,compress]{natbib}
\usepackage{soul}

\usepackage[normalem]{ulem}
\usepackage{tabularx,ragged2e,booktabs}
\usepackage{enumitem}

\usepackage[outline]{contour}
\usepackage{mathtools}
\usepackage{bbm}

\makeatletter
\renewcommand{\paragraph}{%
  \@startsection{paragraph}{4}%
  {\z@}{1.05ex \@plus 1ex \@minus .2ex}{-1em}%
  {\normalfont\normalsize\bfseries}%
}
\makeatother

\def\wacvPaperID{1823} 

\wacvalgorithmstrack   

\wacvfinalcopy 

\ifwacvfinal
\usepackage[breaklinks=true,bookmarks=false]{hyperref}
\else
\usepackage[pagebackref=true,breaklinks=true,colorlinks,bookmarks=false]{hyperref}
\fi

\usepackage{enumitem,amssymb}
\newlist{todolist}{itemize}{2}
\setlist[todolist]{label=$\square$}
\usepackage{pifont}

\definecolor{_fbteal3}{HTML}{CBCBCB} 
\definecolor{Gray}{gray}{0.85}
\newcolumntype{a}{>{\columncolor{_fbteal3}}c}
\newcommand{\putalg}{{PAWS}\xspace}
\newcommand{\putouralg}{\textsc{lava}\xspace}

\newcommand{\xv}{\boldsymbol{x}}
\newcommand{\yv}{\boldsymbol{y}}
\newcommand{\zv}{\boldsymbol{z}}
\newcommand{\uv}{\boldsymbol{u}}
\newcommand{\qv}{\boldsymbol{q}}
\newcommand{\pv}{\boldsymbol{p}}
\newcommand{\mv}{\boldsymbol{m}}

\newcommand{\Lcal}{\mathcal{L}}

\newcommand{\RN}[1]{%
	\textup{\lowercase\expandafter{\it \romannumeral#1}}%
}

\usepackage[normalem]{ulem}
\usepackage{subfig}
\usepackage{colortbl}

\begin{document}

\title{\putouralg: \underline{{\color{blue}La}}bel-efficient \underline{{\color{blue}V}}isual Learning and \underline{{\color{blue}A}}daptation}

\author{Islam Nassar\textsuperscript{1}\thanks{corresponding author: islam.nassar@monash.edu}, Munawar Hayat\textsuperscript{1}, Ehsan Abbasnejad\textsuperscript{2}, Hamid Rezatofighi\textsuperscript{1}, \\ Mehrtash Harandi\textsuperscript{1}, Gholamreza Haffari\textsuperscript{1}\\
{\small \textsuperscript{1} Monash University, Australia \quad \textsuperscript{2} University of Adelaide, Australia}
}


\maketitle
\thispagestyle{empty}

\begin{abstract}

We present \putouralg, a simple yet effective method for multi-domain visual transfer learning with limited data. \putouralg builds on a few recent innovations to enable adapting to partially labelled datasets with class and domain shifts. First, \putouralg learns self-supervised visual representations on the source dataset and ground them using class label semantics to overcome transfer collapse problems associated with supervised pretraining. Secondly, \putouralg maximises the gains from unlabelled target data via a novel method which uses multi-crop augmentations to obtain highly robust pseudo-labels. By combining these ingredients, \putouralg achieves a new state-of-the-art on ImageNet semi-supervised protocol, as well as on 7 out of 10 datasets in multi-domain few-shot learning on the Meta-dataset.\footnote{Code:\url{github.com/islam-nassar/lava.git}}

\end{abstract}

\section{Introduction}
\label{sec:intro}

Using limited data to effectively adapt to new tasks is a challenging but essential requirement for modern deep learning systems. It enables leveraging the power of such systems while avoiding excessive data annotation which is usually costly, time consuming, and often requires domain expertise \cite{caron2021emerging, grill2020bootstrap,wei2020can}. A promising direction is to develop methods that are capable of transferring knowledge across collective data of many tasks. In this work, we examine low-resource multi-domain visual transfer: given a visual learner pretrained on a source dataset\footnote{We use ImageNet~\cite{russakovsky2015imagenet_large} as the source dataset in all our experiments.}, our goal is to effectively transfer to a target dataset with potential class and/or domain shift. We focus on low-resource cases where the target dataset is very small but fully labelled (as in few-shot learning - FSL); or cases where it is sufficiently large but only partially labelled (as in semi-supervised learning - SSL). We propose  a transfer method which simultaneously addresses such cases and demonstrates a strong performance on multiple transfer benchmarks.

\begin{figure}[t]
    \includegraphics[width=1\linewidth, trim={1.2cm 0 0 0}]{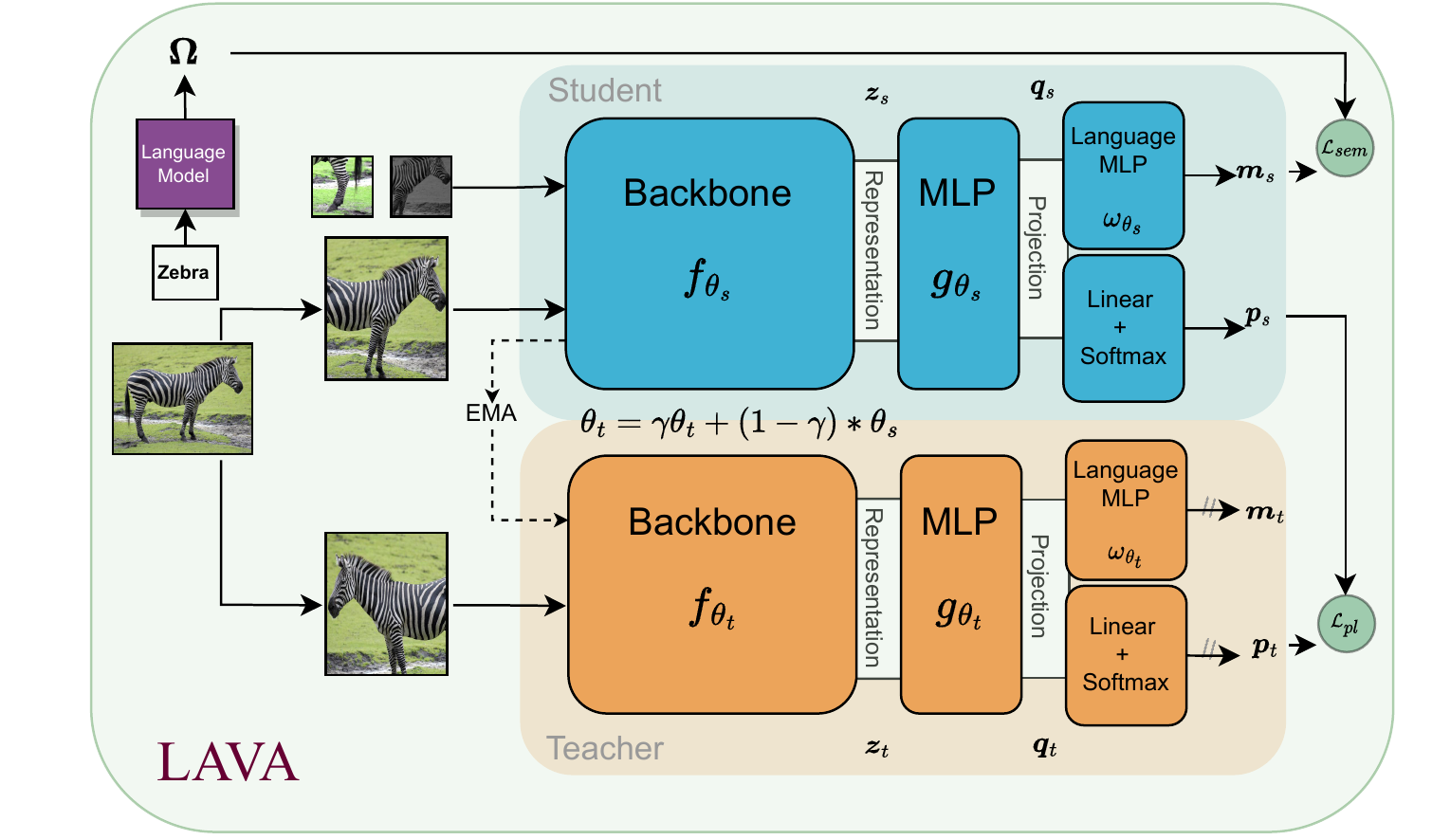}
  \centering
  \caption{\small{\textbf{Method Overview.} \putouralg uses the source dataset to learn self-supervised initialisations as well as a mapping between visual features and label semantics. During transfer, \putouralg uses a teacher-student setup to adapt by multi-crop pseudo-labeling unlabelled instances, and by matching the semantics of labelled ones.}\vspace{-4mm}}
  \label{fig:overview}
\end{figure}

\begin{figure*}
    \centering
    \vspace{-3mm}
    \scalebox{0.96}{\includegraphics[width=1\textwidth]{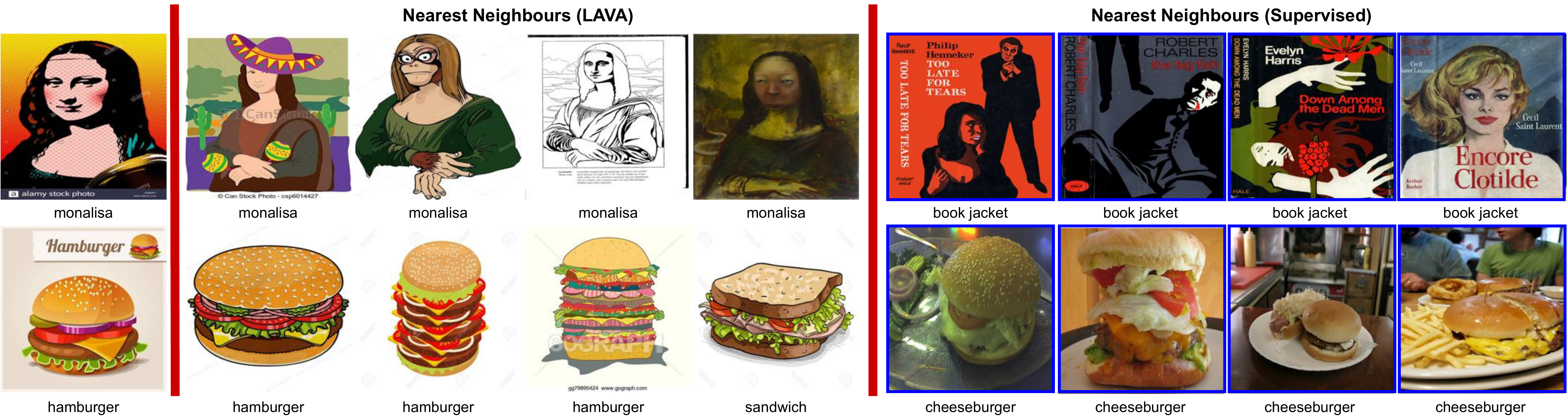}}
    \caption{\small{\textbf{Class \& Domain collapse.} \small We display two query images from clipart target (left), and their 4 nearest neighbours, from clipart and ImageNet, in \putouralg's representation space (middle) and an ImageNet-supervised learner's (right) \emph{before training on target instances}. ImageNet instances are bordered in blue. We observe that the supervised learner suffers from two types of representation collapse: (i) class collapse (upper), where the learner picks an irrelevant source class which shares superficial patterns with target class; or (ii) domain collapse (lower), where the learner picks a relevant source class but ignores the visual domain. Please refer to Sec.~\ref{sec:analysis} for more details.}}
    \label{fig:sup_collapse}
\end{figure*}

\putouralg's first design goal is to employ a pretraining strategy which supports generalisation beyond classes and domains with limited labelled data. Hence, we begin by investigating the effect of source pretraining on the visual transfer performance. In line with recent research, we find that supervised pretraining leads to sub-optimal transfer\cite{doersch2020crosstransformers, assran2021semisupervised, nassar2021all}. Supervision with labels, more often than not, is too eager to learn specialised features which can successfully discriminate the source classes/domains but fails to generalise beyond them. We argue that its other limitation is the semantic-free nature of the labels used. Labels are represented using one-hot vectors explicitly encouraging to ignore label semantics. For example, the learner is encouraged to treat ``bus'' and ``school bus'' as two unrelated classes. In Fig.~\ref{fig:sup_collapse}, we use DomainNet~\cite{peng2019moment} clipart dataset to qualitatively demonstrate two artifacts associated with supervised pretrained representations: class collapse, whereby the pretrained representations collapse into incorrect source classes just because they share superficial similarities with target classes; and domain collapse, where the class semantics are preserved but the visual domain information is disregarded. \putouralg's first ingredient is introducing a two-fold approach to address the collapse problem: 1) self-supervised source pretraining to learn task-independent features leading to better transfer; and 2) using the language modality to ground the self-supervised representations to an independent semantic space: during pretraining, \putouralg uses source class labels to learn a mapping between the visual representations of the source instances and the language semantic representation of their class labels. At transfer time, such mapping is used to infer relations between ``seen'' and ``unseen'' classes by virtue of their foreknown semantic similarities. (Sec.~\ref{subsec:generalising})

\begin{figure*}[t]
    \centering
    \includegraphics[width=\textwidth]{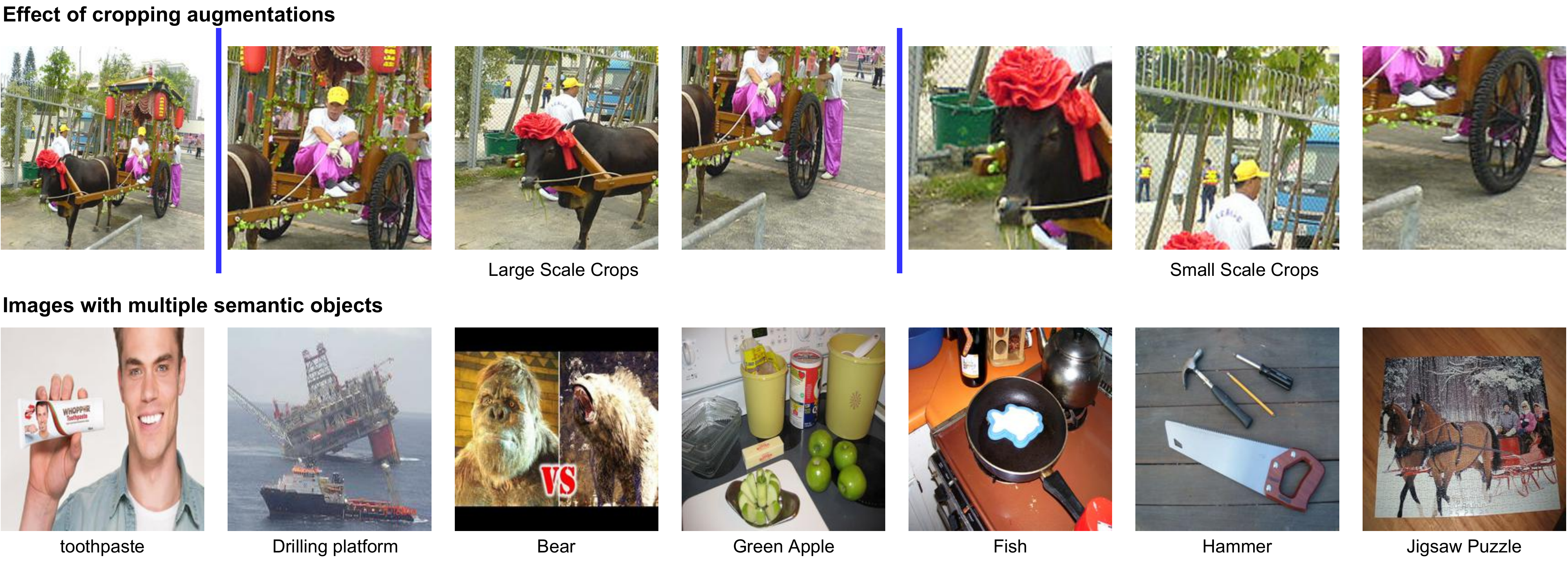}
    \caption[Caption for LOF]{We display the effect of cropping (top), commonly used in vision training, to highlight our multi-crop pseudo-labelling motivation: cropping can potentially change the main focus to a different semantic object. Since in a teacher-student setup, each network receives a different view of the image, images with multiple semantic objects (bottom)~\footnotemark can hurt performance if we rely on a single pseudo-label per image.}
    \label{fig:motivation}
\end{figure*}

\putouralg's second design goal is to leverage unlabelled target data (if available) to improve transfer performance. For that, we employ multi-crop augmentation which was originally proposed to encourage learning representations that are invariant to spatial augmentations for self-supervised contrastive learning~\cite{caron2020unsupervised}. We extend the idea to semi-supervised learning exploiting the observation that images can often contain multiple semantic classes and hence using a single label per image can hurt the performance. Consider the image in Fig.~\ref{fig:motivation}, due to random cropping used during training, it is conceivable that the teacher model (performing the pseudo-labeling) receives a view centered around a different object than the student, leading to a label that is not compatible with the image. Therefore, we calculate pseudo-labels based on multiple local and global views of images to account for those containing multiple semantic concepts. We show that enforcing a single pseudo-label per image is sub-optimal. Instead, applying a pair-wise aggregate loss across multiple views enhances the quality of pseudo-labels. (Sec.~\ref{subsec:method_multi_crop_pseudo_label})

\putouralg's main contributions are:\textbf{1)} a practical method which combines and extends a few recent innovations to address various transfer learning regimes; \textbf{2)} provides empirical insights about transfer collapse problems associated with supervised pretraining and proposes a strategy to address them; \textbf{3)} extends the multi-crop augmentation strategy to the semi-supervised setting via pseudo-labeling; \textbf{4)} sets a new state-of-the-art on the ImageNet~\cite{russakovsky2015imagenet_large} SSL benchmark and demonstrates strong performance on other challenging SSL and FSL benchmarks including the Meta-Dataset~\cite{triantafillou2019meta}.

\section{Related Work}
\label{sec:related_work}

\noindent \textbf{Few-Shot learning (FSL)}
 Existing FSL approaches can be categorized into metric-learning methods that learn an embedding space to compare query and support samples~\cite{snell2017prototypical, koch2015siamese, vinyals2016matching, oreshkin2018tadam}, meta-learning approaches that adapt a base learner to new classes \cite{finn2017model, li2017meta, jamal2019task, rusu2018meta, bertinetto2018meta, lee2019meta, ravichandran2019few}, or a combination of both \cite{triantafillou2019meta}. Most existing FSL methods work well when the train and test sets are from the same domain \eg subsets of ImageNet (mini-ImageNet \cite{vinyals2016matching}, and tiered-ImageNet \cite{ren2018meta}). They lack  out-of-domain (OOD) generalization, once there is a distribution mismatch between train and test data. 
Recently introduced meta-dataset \cite{triantafillou2019meta} provides a challenging benchmark for multi-domain FSL.
FSL methods developed on meta-dataset therefore aim to tackle OOD generalization \cite{liu2020universal, doersch2020crosstransformers}. For instance, Transformer module is employed to capture relationships between different domains in \cite{liu2020universal, doersch2020crosstransformers}. 
Even though our approach is generic across different low-label regimes, our results on meta-dataset show that we perform favorably against recent approaches which are specifically developed for FSL.

\smallskip
\noindent  \textbf{Semi-supervised learning}.
A common approach for SSL is to train the model with a joint loss formulation \ie a supervised cross-entropy loss for the labelled samples, and a un/self-supervised regularization term for unlabelled samples.  Examples include UDA \cite{xie2020unsupervised}, S4L \cite{zhai2019s4l} and  \cite{grandvalet2004semi, miyato2018virtual, verma2021interpolation}.  
Another approach to SSL is using pseudo-labels which are generated by either training the  model on the labelled samples and pruning the confident predictions on unlabelled data \cite{berthelot2019mixmatch,berthelot2019remixmatch,sohn2020fixmatch, nassar2021all, lee2013pseudo, riloff1996automatically}, or by using a teacher-student configuration, where a slowly updating teacher model is used to generate soft-predictions on the unlabelled samples, which serve as a supervisory signal for the student model \cite{pham2021meta,xie2020self,tarvainen2017mean}. \putouralg leverages the latter paradigm but improves pseudo-labels using a multi-crop augmentation strategy.

\smallskip
\noindent  \textbf{Semantics and Self-supervision for FSL}.
Rich semantics~\cite{frome2013devise, afham2021rich, radford2021learning} and self-supervision \cite{doersch2020crosstransformers, khosla2020supervised} have been explored to help FSL. ~\cite{afham2021rich} introduces an auxiliary task to produce class-level semantic descriptions and showed improvements on fine-grained tasks,  while ~\cite{radford2021learning} learns rich features from large number of (image, text) pairs. On the other hand, recent work~\cite{tian2020rethinking, doersch2020crosstransformers} recognised the usefulness of self-supervised features and their ability to generalise, ~\cite{khosla2020supervised} adjusted the instance discrimination SimCLR~\cite{chen2020simple} method to use the few support labels for positive and negative mining, while \cite{doersch2020crosstransformers} explicitly adds SimCLR \cite{chen2020simple} episodes to the training pipeline. Unlike existing approaches, \putouralg only uses class-level label semantics and employs self-supervision~\cite{caron2021emerging} as a pretraining step rather than incorporating it into the FSL task.

\footnotetext{\scriptsize These images are not cherry-picked. They were identified in our analysis (\textit{see} Sec.~\ref{sec:analysis}) to have inconsistent pseudo-labels. }



\section{Method}
\label{sec:method}
We consider the problem of adapting a classifier pretrained on a set of source classes $C_{src}$ using labelled samples $\mathcal{D}_{src}$ to a target dataset of $C_{tgt}$ classes by using labelled instances $\mathcal{D}_{tgt} = \{(\xv^i, \yv^i)\}_{i=1}^n$, and unlabelled instances $\mathcal{U}_{tgt} = \{\uv^j\}_{j=1}^{m}$, with $\uv$, $\xv$ denoting an unlabelled, and labelled image respectively, and $\yv^i$ is the class label. Note that such setup suits both SSL and FSL settings. However, in FSL, $C_{src}$ and $C_{tgt}$ are strictly disjoint and the few-shot transfer utilises a fully labelled support set. 

\putouralg employs a teacher-student setup with a teacher identical in architecture to an online student (\textit{see} Fig.~\ref{fig:overview}). The student is trained to match the ``soft'' label generated by the teacher when each receives different views of a given image. The student and teacher networks, parametrized by $\theta_s$ and $\theta_t$ respectively, are updated alternatively: given a fixed teacher, the student is first updated by gradient descent to minimize the network loss; subsequently, the teacher parameters are updated as an exponential moving average (EMA) of the student's, \ie $\theta_t  \leftarrow \gamma \theta_t + (1 - \gamma) \theta_s$, where $\gamma$ is the momentum parameter. 


\subsection{Generalising beyond domains and classes}
\label{subsec:generalising}
Our first aim is to use the source dataset to pretrain our teacher and student with good initial representations to support out-of-distribution transfer while avoiding the collapse problems mentioned in Sec.~\ref{sec:intro}.  

\smallskip
\noindent \textbf{Self-supervised Pretraining. } We employ the recently proposed DINO~\cite{caron2021emerging} method (without modification) to learn self-supervised initialisations from $\mathcal{D}_{src}$ after discarding the labels. DINO, like other self-supervised methods~\cite{he2020momentum, dosovitskiy2015discriminative,chen2020simple,hjelm2018learning,bachman2019learning, li2021efficient}, learn visual features which are invariant to common factors of variation (\eg colour distortion, pose, scale) while not being tied to a specific set of classes or visual domains. Therefore, they encode richer information which better supports generalisation. 
At transfer time, we use the target instances without their labels to further fine-tune the DINO representations to the target dataset. We provide in the appendix a detailed procedure of fine-tuning DINO features to target instances in low-resource cases. We note that our method is not specifically tied to DINO. However, we chose it due to its demonstrated performance and its similar teacher-student setup making it seamless to integrate with our method. We ablate such choice in our experiments.

\smallskip
\noindent \textbf{Semantic Grounding. }To combat the class collapse problem and to aid generalisation to unseen classes, we employ language semantics as an independent modality to ground visual features. During the source pretraining, we additionally learn a semantic MLP module $\omega_{\theta_*}$ (* denotes s and t for student and teacher respectively) mapping the projection $\qv^i_*$ for a given labelled image $\xv^i$ to an embedding $\mv^i_* =  \omega_{\theta_*}(\qv^i_*)$. We apply a hinge loss which minimizes the discrepancy between the semantic projection $\mv^i_s$~\footnote{\scriptsize We use the student semantic projection $\mv^i_s$ to apply the loss, while $\mv^i_t$ is only used during inference.} and the corresponding embedding $\Omega^i$ obtained by applying a pretrained language model on the class label, as per:

\begin{align}\label{eq:loss_vl}
\Lcal_{sem} = \frac{1}{| \mathcal{D}_{src}|} \sum_{\xv^i \in \mathcal{D}_{src}}&{\max(0, \eta  - \langle \mv^i_s, \Omega^i \rangle + \langle \mv^i_s, \hat{\Omega^i}\rangle)},
\end{align}
 \vspace{-4mm}
 
\noindent where $\langle \cdot, \cdot \rangle$ denotes cosine similarity, $\eta > 0$ is a scalar hinge margin, and $\Omega^i$ and $\hat{\Omega^i}$ are the language embeddings of the true and a randomly-sampled false class respectively. In effect, \putouralg learns how to map the visual representations of the source instances to the language model representation space so that each instance is mapped closer to its true class language embedding and further away (up to a margin) from all other class embeddings. At transfer time, $\omega_{\theta_*}$ is fine-tuned (together with the backbones) without re-initialisation. This is in contrast to the classifier head which must be re-initialised to match the target classes. 

Unlike the discrete nature of one-hot class labels, using language embeddings to ground visual representations acts as a continuous space to represent class labels. In such space, ``bench'' is close to ``park bench'' and ``zebra'' is closer to ``horse'' than it is to ``car''. This is intuitively useful to learn visual-semantic relations which enhance generalisation to novel concepts. However, it also implicitly assumes that linguistic similarity is always a good proxy for visual similarity, which is sometimes not true \eg ``wine glass'' and ``wine bottle'' while linguistically similar, are usually visually distinct. Hence, we explored few alternatives for the source of semantics including knowledge graph embeddings~\cite{nassar2021all}, ``Glove''~\cite{pennington2014glove} and other variants of language models (\textit{see} appendix for details). We find that language models trained on paraphrasing tasks~\cite{song2020mpnet} provide the best performance in our setup. We conjecture that this is because it helps the model to unify similar visual concepts which appear under different names across datasets (\eg ``airplane'' vs ``plane'', ``horse cart'' vs ``carriage'', \etc).

\subsection{Multi-crop pseudo-labelling}
\label{subsec:method_multi_crop_pseudo_label}
When transferring to a partially labelled dataset, \putouralg leverages unlabelled samples by using the teacher to iteratively produce pseudo-labels to expand the labelled samples used to train the student. We differ from previous similar SSL approaches (\eg MeanTeacher~\cite{tarvainen2017mean}) primarily in the way we generate the pseudo-labels: we encourage more robust pseudo-labels via multi-crop augmentation. 
We generate pseudo-labels based on a set $\mathcal{T}^i$ of multiple sized crops of a given unlabelled image $\uv^i$; similarly, student predictions are generated based on another set $\mathcal{S}^i$\footnote{\scriptsize Spatial augmentations (\eg color jittering, random flips) are also applied on all the crops.}. The pseudo-label is then aggregated over the combined views.  

Formally, \putouralg uses the backbone $f_{\theta_*}$ to map an unlabelled image $\uv^i$ to $\zv^i_* = f_{\theta_*}(\uv^i)$, followed by an MLP $g_{\theta_*}$ mapping $\zv^i_*$ to a projection $\qv^i_* = g_{\theta_*}(\zv^i_*)$. Finally, using a linear layer followed by a temperature sharpened softmax, $\qv^i_*$ is normalised into a probability distribution $\pv^i_* \in \mathbb{R}^{||C_{tgt}||}$. Then, we apply our loss as per:
$\Lcal_{pl} = \frac{1}{| \mathcal{U}|}\sum_{\uv^i \in \mathcal{U} } \Lcal_{multi}^i$, where $\Lcal_{multi}^i$ represents the aggregate loss over targets $\pv^i_t | \uv^i \in \mathcal{T}^i$ and predictions $\pv^i_s | \uv^i \in \mathcal{S}^i$.

\smallskip
\noindent \textbf{Design choices. }Aggregating multi-crop losses involves a few design alternatives such as: using a single pseudo-label (\eg via voting) or averaging across the different crops;  
using hard pseudo-labels (\eg using argmax or sampling) or soft pseudo-labels; and finally, the count, scale, and size of the crops are important hyperparameters as they respectively impact the diversity of pseudo-labels, the locality of the crops, and the memory consumption during training. Our empirical study found that using soft sharpened pseudo-labels and averaging over pairs of crops yields the best performance across different domains (refer to appendix for more details). More concretely, we eventually opted for: $\Lcal_{multi}^i = \frac{1}{| \mathcal{A}^i|} \sum_{ (\Tilde{\uv^j}_s, \Tilde{\uv^j}_t) \in \mathcal{A}^i } -  \pv^j_{t} \log \pv^j_s,$
where 
($\Tilde{\uv}^j_s$, $\Tilde{\uv}^j_t$) is a pair of crops of $\uv^i$ passed to the student and teacher respectively; and $\mathcal{A}^i = \{(\Tilde{\uv}^j_s,\Tilde{\uv}^j_t)|\Tilde{\uv}^j_s \in \mathcal{S}^i, \Tilde{\uv}^j_t \in \mathcal{T}^i\}$ is the set of all crop pairs. 
\begin{table*}[h!]
\centering
\caption{\small{Comparison to semi-supervised baselines on four domains of the DomainNet dataset. We report average accuracy over 3 runs for three amounts of labelled instances. All the baselines are implemented in the same codebase using the same network backbone.}}
\label{tab:domainnet_semisup}
\scalebox{0.8}{
\begin{tabular}{llccclccclccclccc}
\toprule[0.15em]
 &  & \multicolumn{3}{c}{\bf Real} &  & \multicolumn{3}{c}{\bf Clipart} &  & \multicolumn{3}{c}{\bf Sketch} &  & \multicolumn{3}{c}{\bf Quickdraw} \\ \cline{3-5} \cline{7-9} \cline{11-13} \cline{15-17} 
 &  & 2-shot & 4-shot & 8-shot &  & 2-shot & 4-shot & 8-shot &  & 2-shot & 4-shot & 8-shot &  & 2-shot & 4-shot & 8-shot \\ \midrule[0.15em]
FixMatch~\cite{sohn2020fixmatch} &  & 23.06 & 34.68 & 42.14 &  & 30.21 & 41.21 & 51.29 &  & 12.73 & 21.65 & 33.07 &  & 24.51 & 32.98 & 43.91 \\
SemCo~\cite{nassar2021all} &  & 24.38 & 40.03 & 51.13 &  & 28.39 & 46.96 & 55.48 &  & 15.71 & 28.62 & 41.06 &  & 26.17 & 34.17 & 44.12 \\
MeanTeacher++~\cite{tarvainen2017mean} &  & 51.44 & 66.16 & 68.77 &  & 46.02 & 52.43 & 63.09 &  & 25.8 & 38.79 & 51.16 &  & 29.78 & 39.12 & 47.11 \\ \midrule[0.15em]
\rowcolor{_fbteal3}
\putouralg (supervised init.) &  & 57.47 & \textbf{69.51} & \textbf{75.41} &  & 38.45 & 53.05 & 64.74 &  & 36.15 & 45.15 & 52.15 &  & 32.61 & 41.67 & 48.44 \\
\rowcolor{_fbteal3}
\putouralg (no semantic loss) &  & 58.57 & 67.88 & 72.12 &  & 48.57 & 58.75 & \textbf{65.18} &  & 38.76 & 47.55 & 53.91 &  & 35.95 & 44.01 & \textbf{54.91} \\
\rowcolor{_fbteal3}
\putouralg &  & \textbf{58.79} & 68.04 & 72.19 &  & \textbf{48.65} & \textbf{59.05} & 65.08 &  & \textbf{39.12} & \textbf{47.63} & \textbf{54.39} &  & \textbf{36.66} & \textbf{44.12} & 54.75 \\ \bottomrule[0.15em]
\end{tabular}%
}

\end{table*}

\section{Experiments}
\label{sec:experiment}

We evaluate \putouralg against state-of-the-art (SOTA) baselines in three regimes: \textbf{1)} SSL transfer with domain shift on DomainNet~\cite{peng2019moment}; \textbf{2)}  SSL without domain shift on ImageNet~\cite{russakovsky2015imagenet_large}, and \textbf{3)} multi-domain FSL on Meta-dataset~\cite{triantafillou2019meta}.

\smallskip
\noindent \textbf{Training. }Unless otherwise stated, we use a batch size of 256 with a learning rate of $5e^{-4}$ and Adam~\cite{kingma2014adam} optimizer with a cosine scheduler.  For our multi-crop pseudo-labelling, we use $6$ small scale crops and $2$ large scale crops (different for teacher than student) following same scales in~\cite{caron2021emerging}, and a teacher momentum $\gamma = 0.996$. We use {\textit {mpnet-base-v2}}~\cite{song2020mpnet} language model\footnote{\scriptsize \url{github.com/UKPLab/sentence-transformers}} to obtain the label embeddings for our semantic loss. We report accuracy based on the softmax classifier (\textit{see} Fig.~\ref{fig:overview}), but when relevant, we compare it with the K-Nearest Neighbour (KNN with K=20) accuracy based on the representation $\zv_t$ and/or the semantic accuracy obtained by applying a cosine classifier on the semantic projection $\mv_t$. 

\smallskip
\noindent \textbf{SSL on DomainNet.} This dataset includes images from 6 visual domains spanning 345 object classes. We examine \putouralg's ability to transfer from ImageNet to 4 domains with decreasing similarity: \textit{real}, \textit{clipart}, \textit{sketch}, and \textit{quickdraw}. To ensure fixed settings across all baselines (\eg labelled splits, backbone, learning rate schedule, \etc), we follow recommendations in~\cite{oliver2018realistic} and re-implement the three closest baselines in our codebase. FixMatch~\cite{sohn2020fixmatch} uses consistency regularization with confidence-based pseudo-labelling, SemCo~\cite{nassar2021all} builds on FixMatch but leverages label semantics (via a knowledge graph) to account for pre-known similarities among classes, and MeanTeacher~\cite{tarvainen2017mean} uses momentum teachers for SSL. We extend it to MeanTeacher++ where we employ the same spatial augmentations we use (instead of the original gaussian noise) and we use the same backbone as ours (ViT-S/16~\cite{touvron2021training}). For all experiments, we initialise models with pretrained ImageNet weights and follow the SSL standard approach: we use a fraction of the labelled data (expressed in images/class) together with all the unlabelled data to adapt to target. We fix the training to 70 epochs~\footnote{\scriptsize The 70 epochs are split into 50 epochs of DINO target pretraining and 20 epochs for training \putouralg} (w.r.t unlabelled data) for \putouralg and we use early stopping for all the baseline methods using a validation set. For each of the 4 domains, we examine low-, and moderate-shot transfer scenarios using 2, 4 and 8 images/class. We explore both self-supervised initialisation (\ie DINO) and supervised initialisation~\cite{touvron2021training} for all the baselines. We report the best results among the two for the baselines and both results for \putouralg in Tab.~\ref{tab:domainnet_semisup}. Finally, to examine the contribution of our semantic loss to SSL, we report results for \putouralg when switching it off.

We observe \putouralg outperforms baselines consistently, sometimes, with large margins. Interestingly, FixMatch and SemCo obtains their best results using supervised source pretraining rather than self-supervised (\textit{see} self-supervised results in appendix). One possible explanation is that this is due to the very different method of augmentations used in DINO pretraining compared to FixMatch and SemCo. As conjectured, we see that the impact of self-supervised initialisation for \putouralg becomes more significant when the visual domain is different from that of ImageNet. For example, we observe an impressive 10\% boost in the clipart 2-shot setting from 38.4\% to 48.6\%, proving that self-supervised features helps generalisation beyond domains. 
Among the baselines, MeanTeacher++ is the closest to \putouralg; the main two differences are our multi-crop pseudo-labelling strategy and the semantic loss. We witness a significant boost of \putouralg over MeanTeacher especially with fewer labelled samples. This confirms the usefulness of our multi-crop pseudo-labelling strategy in low-data regimes. Finally, we get a marginal boost when using semantic loss in SSL across almost all experiments. 

\smallskip
\noindent \textbf{SSL on ImageNet.} To examine SSL transfer under the same domain, we follow ImageNet evaluation protocol by using 1\% and 10\% of the labels to train \putouralg. Due to the demanding computational requirements for running experiments on ImageNet, we opted to only re-run SOTA method PAWS~\cite{assran2021semisupervised} with the same ViT-S backbone as ours. PAWS combines self-supervised learning with a non-parametric method of generating pseudo-labels based on a small labelled set. For all the other baselines, unless ViT-S results are reported in the original work, we only report the Resnet50 results. We note however, that the ViT-S model has less parameters (21M) compared to Resnet50 (24M), yet recent work~\cite{touvron2021training, caron2021emerging} showed an approximate 1-2\% improvement in favour of ViT-S. 
Again, we see (Tab.~\ref{table:imagenet_results}) a significant boost for \putouralg against other methods. Interestingly, as opposed to DINO, \putouralg achieves large gains. Note, however, that DINO reports Linear evaluation results on frozen features and does not fine-tune end-to-end, so (64.5\% and 72.2\%) are not directly comparable to the (69.3\% and 76.4\%) of \putouralg. However, K-NN results can be directly compared to measure the differential between \putouralg and DINO. 

\begin{table}[t]
\centering
\caption{\small SSL results on ImageNet with 1\% \& 10\% of the labels.}
\label{table:imagenet_results}
\setlength{\tabcolsep}{2pt}
\scalebox{0.9}{
    \begin{tabular}{l r c c c}
        \small Method & \small Arch. & \small Epochs & \small 1\% & \small 10\% \\\toprule [0.15em]
        \multicolumn{4}{l}{\footnotesize\itshape Different Architecture:}\\[1mm]
        FixMatch~\cite{sohn2020fixmatch} & RN50 & 300 & -- & 71.5 \\
        MPL~\cite{pham2021meta} & RN50 & 800 & -- & 73.9 \\
        SwAV~\cite{caron2020unsupervised} & RN50 & 800 & 53.9 & 70.2 \\
        SimCLRv2++ ~\cite{chen2020big} & RN50 & 1200 & 60.0 & 70.5 \\\midrule[0.15em]
        \multicolumn{4}{l}{\footnotesize\itshape Same Architecture:}\\[1mm]
        DINO-NN~\cite{caron2021emerging} & ViT-S & 300 & 61.3 & 69.1 \\
        DINO~\cite{caron2021emerging} & ViT-S & 350 & 64.5 & 72.2 \\
        \putalg-\textsc{nn}~\cite{assran2021semisupervised} & ViT-S &  300 &  63.5 &  72.3 \\
        \putalg~\cite{assran2021semisupervised} & VIT-S &  300 &  68.9 &  75.2 \\
        \midrule[0.15em]
        \rowcolor{_fbteal3}
        \putouralg-\textsc{nn} & ViT-S & 300 & \bf 67.2 & \bf 73.3 \\
        \rowcolor{_fbteal3}
        \putouralg & ViT-S & 350 & \bf 69.3 & \bf 76.4 \\
        \bottomrule[0.15em]
    \end{tabular}}
\vspace*{-1.5em}
\end{table} 
\begin{table*}[ht]
	\centering
		\caption{Results on Meta-Dataset when only pre-training on ImageNet train split. Mean accuracy, 95\% confidence interval are reported over 600 test episodes. Our method outperforms the best method in 7 out of 10 datasets.}
		\label{tab:meta_dataset_results}
    \resizebox{1.0\textwidth}{!}
    {
		\begin{tabular}{lccccccccca}

		    \toprule [0.2em]
		     & K-NN & MatchNet & ProtoNet & Finetune & fo-Proto & BOHB & ProtoNet-L& ALFA-fo-Proto & CTX & \putouralg \\
		     & \cite{triantafillou2019meta} & \cite{triantafillou2019meta} & \cite{triantafillou2019meta} &  \cite{triantafillou2019meta} &MAML  ~\cite{triantafillou2019meta} &  \cite{saikia2020optimized} &  \cite{triantafillou2019meta} &  MAML\cite{baik2020meta} & \cite{doersch2020crosstransformers} & \xspace(ours) ~ \\		    
		    \midrule [0.15em]
		    ImageNet & 41.03$\pm$1.01 & 45.00$\pm$1.10 & 50.50$\pm$1.08 & 45.78$\pm$1.10 & 49.53$\pm$1.05 & 51.92$\pm$1.05 & 53.69$\pm$1.07 & 52.80$\pm$1.11 & 62.76$\pm$0.99 & {\bf 68.75$\pm$0.54}   \\
		    Omniglot & 37.07$\pm$1.15 & 52.27$\pm$1.28 & 59.98$\pm$1.35 &60.85$\pm$1.58 & 63.37$\pm$1.33 & 67.57$\pm$1.21 & 68.50$\pm$1.27 & 61.87$\pm$1.51 & {\bf 82.21$\pm$1.00} & 77.92$\pm$0.50 \\
		    Aircraft &  46.81$\pm$0.89 & 48.97$\pm$0.93 & 53.10$\pm$1.00 & 68.69$\pm$1.26 & 55.95$\pm$0.99 & 54.12$\pm$0.90 & 58.04$\pm$0.96 & 63.43$\pm$1.10 & 79.49$\pm$0.89 & {\bf 81.14$\pm$0.49}\\
		    Birds & 50.13$\pm$1.00 & 62.21$\pm$0.95 & 68.79$\pm$1.01 & 57.31$\pm$1.26 & 68.66$\pm$0.96 & 70.69$\pm$0.90 & 74.07$\pm$0.92 & 69.75$\pm$1.05 & 80.63$\pm$0.88 & {\bf 84.88$\pm$0.77}\\
		    Textures &  66.36$\pm$0.75 & 64.15$\pm$0.85 & 66.56$\pm$0.83 & 69.05$\pm$0.90 & 66.49$\pm$0.83 & 68.34$\pm$0.76 & 68.76$\pm$0.77 & 70.78$\pm$0.88 & 75.57$\pm$0.64 & {\bf 82.05$\pm$0.50} \\
		    Quick-Draw & 32.06$\pm$1.08 & 42.87$\pm$1.09 &48.96$\pm$1.08 & 42.60$\pm$1.17 & 51.52$\pm$1.00 & 50.33$\pm$1.04 & 53.30$\pm$1.06 & 59.17$\pm$1.16 & {\bf 72.68$\pm$0.82} & 68.44$\pm$0.57\\
		    Fungi & 36.16$\pm$1.02 & 33.97$\pm$1.00 & 39.71$\pm$1.11 & 38.20$\pm$1.02 & 39.96$\pm$1.14 & 41.38$\pm$1.12 & 40.73$\pm$1.15 & 41.49$\pm$1.17 & 51.58$\pm$1.11 & {\bf 55.02$\pm$0.67}\\
		    Flowers &  83.10$\pm$0.68 & 80.13$\pm$0.71 & 85.27$\pm$0.77 & 85.51$\pm$0.68 & 87.15$\pm$0.69 & 87.34$\pm$0.59 & 86.96$\pm$0.73 & 85.96$\pm$0.77 & {\bf 95.34$\pm$0.37} & {\bf 95.43$\pm$0.66}\\
		    Traffic-Sign & 44.59$\pm$1.19 & 47.80$\pm$1.14 & 47.12$\pm$1.10 & 66.79$\pm$1.31 & 48.83$\pm$1.09 & 51.80$\pm$1.04 & 58.11$\pm$1.05 & 60.78$\pm$1.29 & {\bf 82.65$\pm$0.76} & 69.24$\pm$1.06  \\
		    MSCOCO & 30.38$\pm$0.99 & 34.99$\pm$1.00 & 48.03$\pm$0.99 & 41.00$\pm$1.10 & 34.86$\pm$0.97 & 43.74$\pm$1.12 & 41.70$\pm$1.08 & 48.11$\pm$1.14 & 59.90$\pm$1.02 & {\bf 63.75$\pm$0.45}\\
			\bottomrule [0.15em] \\
		\end{tabular}%
			}
\end{table*}

\smallskip
\noindent \textbf{FSL on Meta-dataset.}. We use the ``ImageNet-only'' protocol to evaluate multi-domain FSL on Meta-dataset \cite{triantafillou2019meta}. Specifically, we use images of 712 out of 1000 classes of ImageNet as our source dataset, choose our hyperparameters by validating on 158 classes and evaluate using episodes coming from the remaining 130 classes of ImageNet as well as other 9 datasets. Meta-dataset measures cross-domain FSL by evaluating on datasets including fine-grained tasks (such as Birds, Aircrafts, Fungi, Flowers, Textures, Traffic Signs), characters and symbols (Omniglot), and real and quickly drawn objects (ImageNet, MSCoCO and Quickdraw). During source pretraining, we use all the instances coming from the 712 train classes without their labels for source pretraining\footnote{\scriptsize Unlike meta-learning methods, we do not use episodes during the training.}. And, we use the same instances with their labels to train the Semantic MLP ($\omega_{\theta}$). During transfer, we freeze the backbone $f_{\theta_s}$ and finetune \putouralg using the support set for 300 epochs. We use the standard Meta-dataset setting where each episode contains varying number of ways (\ie classes) and imbalanced shots (\ie images per class). As per the common practice, we report results averaged over 600 episodes for each dataset. 

As seen in Tab.~\ref{tab:meta_dataset_results}, with such a simple strategy and without resorting to any meta-learning techniques, \putouralg outperforms the closest baseline~\cite{doersch2020crosstransformers} on 7 out of 10 datasets and for the other 3, it exceeds the second best with a large margin. It is important to note here that it is not straight forward to directly compare between the FSL baselines: primarily, because different methods employ different styles of training (\eg meta-training~\cite{snell2017prototypical, finn2017model, doersch2020crosstransformers} vs fine-tuning), different initialisation (self-supervised~\cite{doersch2020crosstransformers} vs supervised~\cite{finn2017model, snell2017prototypical}) and different backbones (Resnet18~\cite{triantafillou2019meta, saikia2020optimized} vs Resnet34~\cite{doersch2020crosstransformers}). However, we think of our method as orthogonal to other methods: since we are using language semantics, which is not available for other methods, we are cautious about directly comparing with them and hence we report the results for reference and not comparison. A possible future direction is to explore how semantics (or other possible ways of grounding class relationships) help other strong methods such as ``Cross Transformers''.





\begin{figure}
\centering
    \includegraphics[trim={0.3cm 0.25cm 0.25cm 3cm},clip, width=0.45\textwidth]{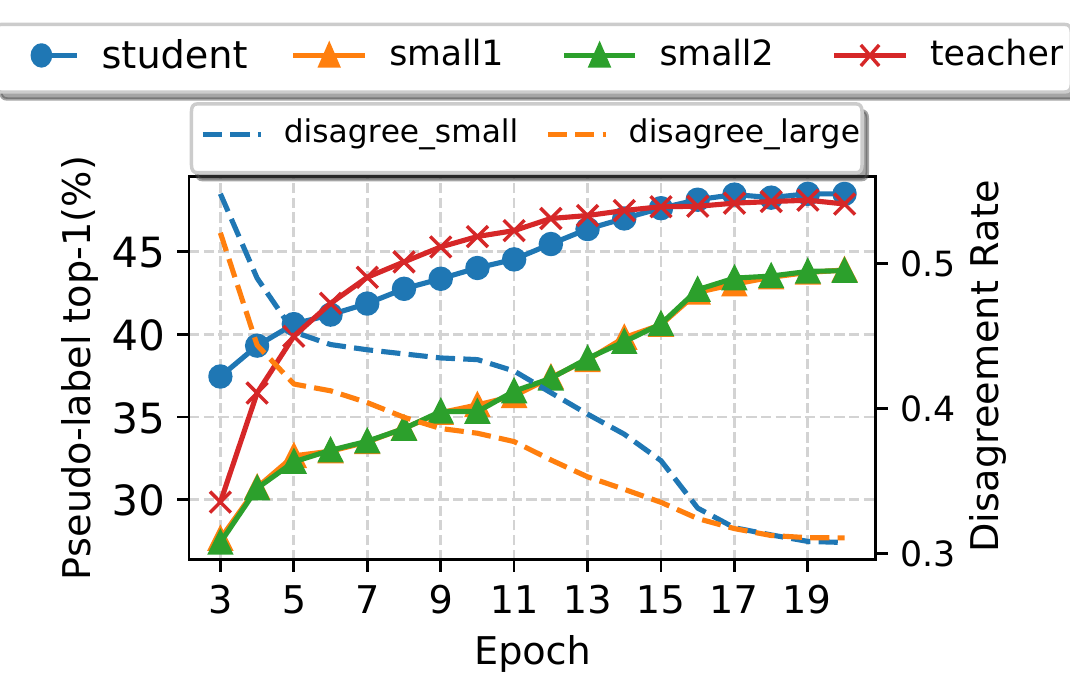}
    \vspace{-2mm}
    \caption{\small Multi-crop Pseudo-labelling analysis.}
\label{fig:multi_crop_analysis}
\end{figure}

\section{Analysis and Ablations}
\label{sec:analysis}
We gain key insights by analysing the following: \textbf{1)} the multi-crop strategy dynamics and its usefulness to produce higher quality pseudo-labels; \textbf{2)} the importance of self-supervised learning for out-of-domain generalisation; \textbf{3)} the role of language semantics towards class generalisation; and finally \textbf{4)} the effect of key hyperparameters. 

\smallskip
\noindent \textbf{Multi-crop Pseudo-labelling. } To study \putouralg's dynamics, we conduct an ``oracle'' experiment where we use the ground truth labels (originally hidden to emulate an SSL setup) to calculate the true pseudo-labels accuracy of the multiple crops seen by the student and the teacher networks as the training proceeds. We calculate accuracies based on the argmax of the soft predictions $\pv_s$ and $\pv_t$ for the student and teacher respectively. Fig.~\ref{fig:multi_crop_analysis} provides few interesting observations. First, based on the large crops, the teacher model exhibits a consistently better performance compared to the student after an initial ramp up phase (due to the EMA). This demonstrates the usefulness of the teacher-student setup, whereby the student model is always guided by a slightly better teacher. Second, as expected, we observe that the small crops (only seen by the student) have lower accuracy, on average, compared to large crops; but interestingly, the disagreement among the predictions associated with the small crops decrease as training proceeds, suggesting that the model is learning from the different small views of an image a consistent representation which truly captures its main object. 
Next, we repeat the same process for our closest baseline (MeanTeacher++) to examine the difference in pseudo-labelling quality obtained by each of the methods. Note that with the modifications we introduce to MeanTeacher++, the difference between the two methods in such setting is only the multi-crop pseudo-labelling. We observe in Tab.~\ref{table:init_study} that indeed, the multi-crop strategy brings a significant benefit across three different domains especially when less labelled data is available. Additionally, we captured a fine-grained view of the pseudo-labels to examine what are the images that most differ in pseudo-labels among the two methods. As expected (\textit{see} Fig.~\ref{fig:motivation} - bottom), those are the images which contain multiple semantic objects. We provide further examples in the appendix.

\begin{table}[t]
\centering
\caption{\small \textbf{Initialisation study.} Cross-domain performance when using different initialisation and different fine-tuning methods. }
\label{table:init_study}
\setlength{\tabcolsep}{2pt}
\scalebox{0.8}{
\begin{tabular}{llcclcclcc}
\toprule[0.15em]
 &  & \multicolumn{2}{c}{\textbf{Real}} &  & \multicolumn{2}{c}{\textbf{Clipart}} &  & \multicolumn{2}{c}{\textbf{Quickdraw}} \\ \cline{3-4} \cline{6-10} 
 &  & 2-shot & 8-shot &  & 2-shot & 8-shot &  & 2-shot & 8-shot \\ \hline
Fully-supervised &  & 73.74 & 76.55 &  & 71.87 & 72.75 &  & 61.22 & 67.23 \\ \midrule[0.1em]
\multicolumn{9}{l}{\itshape Initialisation:}\\[1mm]
Sup. \ \ \ (ImNet)$^*$ &  & 54.92 & 64.81 &  & 22.49 & 35.39 &  & 10.68 & 18.78 \\
DINO (ImNet) &  & 46.17 & 59.42 &  & 15.19 & 26.97 &  & 8.98 & 16.81 \\
DINO (Target)$^{**}$ &  & 50.45 & 62.54 &  & 40.35 & 55.03 &  & 30.05 & 43.64 \\ \midrule[0.1em]
\multicolumn{9}{l}{\itshape Fine-tuning from  *:}\\[1mm]
Linear Probing &  & 49.03 & 64.89 &  & 21.5 & 36 &  & 6.08 & 13.9 \\
MeanTeacher++ &  & 54.26 & 70.60 &  & 34.58 & 60.22 &  & 16.38 & 35.75 \\
\rowcolor{_fbteal3}
\putouralg &  & 57.47 & \bf 75.41 &  & 38.45 & 64.74 &  & 32.61 & 48.44 \\ \midrule[0.1em]
\multicolumn{9}{l}{\itshape Fine-tuning from  **:}\\[1mm]
Linear Probing &  & 49.68 & 64.42 &  & 38.86 &  56.6 &  & 28.73 & 47.5 \\
MeanTeacher++ &  & 51.44 & 68.77 &  & 46.02 & 63.09 &  & 29.78 & 47.11 \\
\rowcolor{_fbteal3}
\putouralg &  & \bf 58.79 & 72.19 &  & \bf 48.65 & \bf 65.08 &  & \bf 36.66 & \bf 54.91 \\ \bottomrule[0.15em]
\end{tabular}}
\vspace*{-1.5em}
\end{table} 

\begin{figure*}
    
    \centering
    \scalebox{0.99}{\includegraphics[trim={0cm 0.5cm 0cm 0.25cm},clip, width=0.92\textwidth]{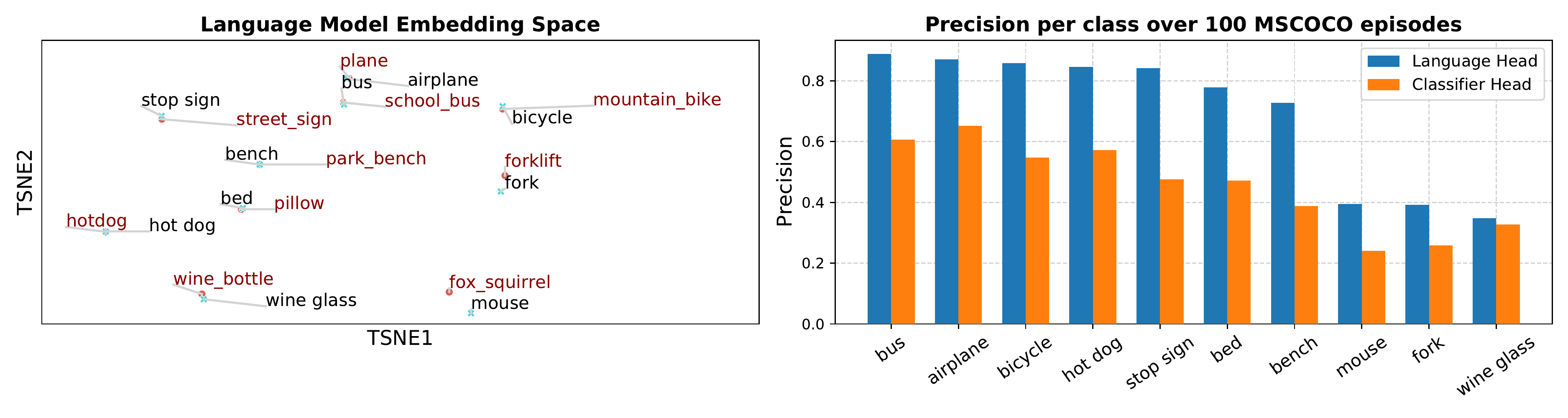}}
    \vspace{-2mm}
    \caption{\small{\textbf{Left}: we display the t-SNE visualization of the language model embeddings for a subset of MSCOCO classes (black) and their nearest neighbours in ImageNet train split (red). \textbf{Right}: we report precision per class over 100 FSL episodes for both the semantic and softmax predictions.}}
    \label{fig:language_vs_precision}
\end{figure*}

\smallskip
\noindent \textbf{Initialisation Study. } Here, we are interested to examine the effect of pretraining when training \putouralg using few labels across different visual domains. In Tab.~\ref{table:init_study}, we report results using different initialisations and different fine-tuning settings across \textit{real}, \textit{clipart}, and \textit{quickdraw} domains. In the top section, we display KNN accuracy based on the model representation $\zv_t$ when initialised with \textbf{1)} Supervised ImageNet features~\cite{touvron2021training}; \textbf{2)} DINO ImageNet features~\cite{caron2021emerging}; and \textbf{3)} DINO features when trained on the target dataset without labels (the standard \putouralg initialisation). Note that for those results, the labels are only used to obtain the KNN accuracy on the validation set of the respective target dataset but never for fine-tuning so they are only meant to compare the quality of ``off-the-shelf'' pretrained features. First, we observe that using only ImageNet data, supervised pretraining is more useful than self-supervised across the three domains and shots with a degrading performance as the domain deviates from ImageNet. Note that 23\% of DomainNet classes also exists in ImageNet, explaining why class-specific features might be helpful in such case. However, once we have access to the target instances (without their labels), we observe that self-supervised target training (\ie \putouralg initialisation) dramatically improves the representations to become more suited to the target domain without any labeling expense. Even when the target domain is very close to ImageNet (\eg \textit{real}), we see a 4\% gain in the 2-shot regime (46.17 to 50.45). This boost is even more pronounced in highly dis-similar domains, \eg 25\% and 21\% boosts for \textit{clipart} and \textit{quickdraw} with only 2 shots per class.

On the other hand, in the middle and bottom sections of Tab.~\ref{table:init_study}, we report transfer results based on the supervised initialisation and the \putouralg initialisation. We also report results using two other transfer methods: 1) Linear Probing on top of frozen representations~\cite{caron2021emerging}; and 2) MeanTeacher++ described in Sec.~\ref{sec:experiment}. Finally, as an upper bound, we report the fully-supervised results obtained when using the entire target dataset (with labels) to train ViT in a supervised manner, then using the few shots to obtain the KNN accuracy reported. Here, we observe that \putouralg benefits from self-supervised DINO initialisation in all domains, but the gain is more clear when the target domain is different than ImageNet and the available labels are less. For example, we witness an impressive 10\%, and 4\% boosts in the 2-shot scenario for \textit{clipart} and \textit{quickdraw}, respectively. Additionally, to quantify what \putouralg brings on top of the DINO initialisation, we compare \putouralg with MeanTeacher++ and observe that \putouralg outperforms it in all cases thanks to our multi-crop pseudo-labelling strategy. Finally, we notice that \putouralg is almost closing the gap to fully-supervised training using all the target labels: by only using 8-shots, \putouralg achieves 75.4\% on the \textit{real} domain compared to the 76.5\% obtained when all the labels are used for training. 

\smallskip
\noindent \textbf{Language Semantics. }Now, we examine the role of label semantics towards generalisation to unseen classes and we investigate if \putouralg indeed mitigates the ``class collapse'' problem. To study the impact of label semantics, we consider the 100 FSL episodes of MSCOCO in the Meta-dataset experiment. Recall that we first pretrain on instances of 712 classes of ImageNet $C_{src}$ then transferred to a different set of classes in MSCOCO $C_{tgt}$. In Fig.~\ref{fig:language_vs_precision}-left, we display a subset of MSCOCO $C \in C_{tgt}$ together with their nearest neighbors among $C_{src}$, when they are projected into the language model semantic space. This space is pretrained using language and so it captures the linguistic semantic similarity between different sentences/words. On the right plot, we display the average precision per class $c \in C$ based on the softmax classification head as well as the semantic projection head\footnote{\scriptsize For a given image, semantic predictions are obtained by finding the class whose language embedding is nearest to $\mv_t$.}. We observe that for any given MSCOCO class, when the most semantically-similar class in ImageNet is also visually similar (\eg ``{\textit bus}'' and ``{\textit {school\_bus}}''), the language head has significantly higher precision per class. In contrast, in cases when the nearest neighbour is not visually similar (\eg  ``{\textit  wine glass}'' and ``{\textit wine bottle}''), both heads have comparable performance. This suggests the benefit of the learnt semantic mapping module $\omega_\theta$: during test time and without any further training, when the model receives an image sharing similar visual features associated with one of the source classes $C_{src}$, the semantic head maps it to the most closely related linguistic concept in $C_{tgt}$. 

\smallskip
\noindent \textbf{Collapse Analysis. }
We follow a similar setup as ~\cite{doersch2020crosstransformers} to investigate \putouralg's ability to avoid transfer collapse (\emph{see} Sec.~\ref{sec:intro}): we begin by uniformly sampling 100 images per class from ImageNet $C_{src}$ as well as 1000 query images from clipart dataset. Subsequently, we compute the representation $\zv_t$ for all of the sampled images. Besides \putouralg, we also compute representations obtained by a supervised learner; and a DINO-initialised learner. In Fig.~\ref{fig:sup_collapse}, we report examples of query images with their 4 nearest neighbours among all the sampled images. The representation of a given query image is said to be collapsed if its nearest neighbours mostly belong to source classes. To quantify the collapse in the three methods, we calculate the percentage of the 10 nearest neighbours which are instances of source classes over the 1000 query images. We find that this figure is 25\% for the supervised learner, 21.7\% for DINO, and 17\% for \putouralg. We provide further details in the appendix.

\smallskip
\noindent \textbf{Hyperparameters sensitivity. }During preliminary experiments, we identified few important design choices to tune: number of large scale and small scale crops, the pseudo-labelling aggregation loss function, the teacher momentum update rate ($\gamma$), and the source of label semantics embeddings.
We used a held-out validation set on each of \textit{real} and \textit{clipart} domains to tune the hyperparameters (except the source of label semantics for which we used MSCOCO validation in the FSL regime), then we obtained a single set of parameters which we used across all experiments in this paper. We refer the reader to the appendix for a complete list of hyperparameters of \putouralg in addition to a study to demonstrate their effect on \putouralg's performance.
\section{Conclusion}
\label{sec:conclusion}

We introduced a unified strategy for multi-domain visual transfer with limited target data. \putouralg employs label semantics and self-supervised pretraining to learn initial representations which support generalisation; and uses multi-crop augmentation to maximise the gains from unlabeled data via pseudo-labeling. We demonstrated \putouralg's success in image classification over multiple benchmarks. We believe our approach is generic and can be extended to other visual learning tasks such as object detection and action recognition. We leave these explorations to future work.

\section*{Acknowledgement}
This work was partly supported by DARPA’s Learning with Less Labels (LwLL) program under agreement FA8750-19-2-0501 and by the Australian Government Research Training Program (RTP) Scholarship.




\clearpage
\section*{Appendix\footnote{\scriptsize All section, table, and figure references are following the original paper.}}
\setcounter{section}{0}
\renewcommand{\thesection}{\Alph{section}}
\section{Training Details}
In this section, we provide more details about our implementation and a full list of the hyperparameters for reproducibility. 

\subsection{Architecture}
We build our system on top of (DINO)'s~\cite{caron2021emerging} github implementation. We use the small Vision Transformer model variant (ViT-S/16) in all our experiments (unless otherwise stated) as it provides the best trade-off between throughput and accuracy and has comparable number of parameters to the baselines with which we compare our method.  However, our implementation also supports the larger model variants which are evaluated in ~\cite{caron2021emerging} (\eg ViT-S/8 , ViT-B/16, and ViT-B/8). 

On top of the ViT-S backbone $\texttt{[CLS]}$ token output representation $\zv_* \in \mathbb{R}^{384}$, we attach our projection MLP $g_{\theta_*}$ (\textit{see} Fig.~\ref{fig:overview}) comprising 3 linear layers with 2048 hidden dimension each followed by a GELU activation except for the last layer which uses an output bottleneck dimension of 256 and a linear activation. Subsequently, we connect our two heads to the output of the MLP $\qv_*$: 1) the language MLP head $\omega_{\theta_*}$ comprises 2 weight-normalized linear layers with GELU activation and a hidden dimension 2048 and output dimension 768; 2) a linear classifier comprising a single weight-normalized linear layer followed by a temperature sharpened softmax (the student network temperature differs from the teacher network temperature). Please refer to Tab.~\ref{tab:hyperparams_pretraining} for a list of hyperparameters.

\subsection{LAVA Training Summary}
\putouralg training comprises a pretraining stage on a source dataset and an adaptation/transfer stage to a target dataset. During source pretraining, \putouralg first learns self-supervised representations by minimizing DINO~\cite{caron2021emerging} loss $\Lcal_{ssl}$. Then it uses the labels to learn a mapping (\ie language MLP) between the frozen self-supervised representations and the label language embeddings using our hinge loss $\Lcal_{sem}$(Eqn. 2). During target training, \putouralg uses a modified variant of the DINO self-supervised objective (described in detail in next section) to adapt its source representations to the target domain without any use of target labels. Subsequently, \putouralg employs a hybrid supervised/unsupervised loss to further train on the target labelled/unlabelled instances: the hinge loss on the labelled instances and our novel multi-crop pseudo-label loss on the unlabelled ones. Importantly, the language MLP is transferred directly from source to target without the need to be reinitialised (as opposed to a linear classifier head in vanilla transfer settings). That is because it predicts a fixed size embedding in a dataset-independent semantic space (the pretrained language model space). Concisely, \putouralg loss can be summarised as per: $\Lcal = \Lcal_{ssl} + \Lcal_{sem} + \Lcal_{pl}$ with modulation coefficients to control the contribution of each term to the different stages of the training pipeline.

\subsection{DINO Target Adaptation}
\label{sec:selfsup_pretraining_details}
After source DINO pretraining, we adapt to target dataset by fine-tuning the source representations on target unlabelled instances using the same DINO objective. In all our experiments, we found that 50 epochs are sufficient to adapt to the target dataset. Empirically, we found that there are two crucial factors to enable the success of this procedure: first, it is important to load the source pretrained weights of the MLP and the DINO head in conjunction with the backbone weights and not just the backbone as commonly done when fine-tuning. The second factor that highly impacts the adaptation procedure is the teacher EMA momentum ($\gamma$) which controls the speed with which the teacher model weights follow the student. We found that allowing the teacher network to update its parameters faster than it did during the source pretraining helps to better adapt to the target dataset. More specifically, instead of using the default value of $\gamma=0.996$ used in ~\cite{caron2021emerging}, we use $\gamma=0.95$ during adaptation. In Tab.~\ref{tab:pretraining_adapt}, we present a comparison between the different options for teacher momentum while adapting to various datasets. We report the KNN accuracy on 3 datasets of DomainNet before adaptation (\ie using ImageNet pretrained model) and after adaptation with two different values for the teacher momentum. We observe that adaptation with lower value of $\gamma$ significantly improves the learnt representations. 

\begin{table}[]
\centering
{%
\scalebox{0.8}{
\begin{tabular}{llll}
\toprule[0.2em]
 & \multicolumn{1}{c}{clipart} & \multicolumn{1}{c}{quickdraw} & \multicolumn{1}{c}{painting} \\ \midrule[0.15em]
KNN before adaptation                   & 51.09 & 35.28 & 56.49 \\
KNN after adaptation ($\gamma = 0.996$) & 65.11 & 58.31 & 59.68 \\
KNN after adaptation ($\gamma = 0.95$)  & 69.08 & 62.29 & 61.58 \\ \bottomrule[0.15em]
\end{tabular}%
}}
\caption{\small KNN validation accuracy for different values of teacher momentum ($\gamma$).}
\label{tab:pretraining_adapt}
\end{table}

\begin{table}[t]
\centering
{%
\scalebox{0.9}{
\begin{tabular}{ll}
\toprule[0.2em]
Hyperparameter                  & Value     \\ \midrule[0.15em]
batch size                      & 256     \\
learning rate                   & 0.0005  \\
optimizer                       & adam \\
minimum learning rate           & 0.000001 \\
warmup learning rate            & True \\
scheduler                       & cosine decay \\
weight decay start/end         & 0.04/0.4 \\
num small crops                 & 8 \\
num large crops                 & 2 \\
global crops scale              & (0.4, 1.0) \\
local crops scale               & (0.05, 0.4) \\
spatial augmentations           & random flip, color jitter, \\&gaussian blur, solarization. \\
out dim (c)                     & 65536 \\
student softmax temperature     & 0.1 \\
teacher softmax temperature     & 0.07 \\
warmup teacher temperature      & True \\
teacher temperature start       & 0.04 \\
teacher momentum start ($\gamma$)         & 0.996/0.95 \\
teacher momentum end            & 1.0 \\\bottomrule[0.15em]
\end{tabular}%
}}
\caption{DINO pretraining default hyperparameters.\vspace{-4mm}}
\label{tab:hyperparams_pretraining}
\end{table}

\begin{table}[]
\centering
{%
\scalebox{0.9}{
\begin{tabular}{ll}
\toprule[0.2em]
Hyperparameter                          & Value     \\ \midrule[0.15em]
learning rate                           & 0.000025  \\
num small crops student                 & 6 \\
num small crops teacher                 & 0 \\
num large crops student                 & 2 \\
num large crops teacher                 & 2 \\
student softmax temperature             & 0.1 \\
teacher softmax temperature             & 0.04 \\
warmup teacher temperature              & False \\
teacher momentum start ($\gamma$)       & 0.99 \\
teacher momentum schedule               & cosine \\
language model                          & {\em mpnet-base-v2} \\
language model latent dimension (d)     & 768 \\
hinge loss margin ($\eta$)              & 0.4 \\\bottomrule[0.15em]
\end{tabular}%
}}
\caption{\putouralg default hyperparameters.\vspace{-4mm}}
\label{tab:hyperparams_finetuning}
\end{table}

\section{Coupling DINO and \putouralg}
We exploit the fact that DINO and \putouralg share a similar self-distillation backbone and only differ in the projection heads architecture, their training procedure and the hyperparameters. Accordingly, to allow training both DINO and \putouralg using the same codebase, we simply attached an additional head to \putouralg's MLP $g_{\theta_*}$ projection with an output space dimension matching that of DINO's (they use 65536 output dimension by default). Subsequently, one can switch between DINO pretraining and \putouralg training by adjusting the running configuration to match the respective setup. An additional benefit for such coupling is that it allows us to add DINO proposed loss as an auxiliary loss to our model objective during adaptation and explore if it introduces any benefits to \putouralg. The intuition is that since we use DINO for \putouralg source pretraining and target fine-tuning, it might be useful to continue applying its loss as an auxiliary loss so as to prevent damaging or ``forgetting'' the learnt self-supervised representations. However, we only found marginal benefits of doing so in the very limited label regimes (\eg 1 and 2-shot experiments in SSL) but such benefits diminish once we have more labelled data.

\section{Ablation Study}
Here, we are interested to examine the effect of key design choices on \putouralg's performance. 
\subsection{Semantics for FSL}

\noindent \textbf{Loss Ablations. }
In Tab.~\ref{tab:domainnet_semisup}, we demonstrated a marginal benefit for our proposed semantic loss under the SSL setting and we concluded that the semantic loss is tailored specifically to address generalisation to new classes. To evaluate such claim, we conduct another experiment in the FSL setting on MSCOCO dataset: starting from \putouralg's base learner, described in Sec.~\ref{sec:experiment}, we run 100 test episodes of MSCOCO (using the same random seed for the episode generation) while ablating over three different losses: standard classification loss using one-hot labels as targets, our proposed semantic loss using language semantics as targets, and our proposed multi-crop pseudo-labelling loss. In Tab.~\ref{tab:loss_ablations_ssl_and_fsl}, we report the ablation results of both SSL and FSL regimes side-by-side for comparison. First, we observe that pseudo-label loss helps in both cases but its role is more evident in SSL as expected. In contrast, we observe that the semantic loss plays an important role in FSL: when combined with pseudo-label loss, it achieves a 10\% boost in accuracy when compared with the standard classification loss (67.68\% vs 57.25\%) while the difference between the same cases in SSL is marginal (48.89\% vs 48.57\%). This confirms the usefulness of rich semantics to generalise to unseen classes as conjectured earlier. Finally, we observe that using semantic and pseudo-labeling losses (\putouralg standard setting), we obtain the best performance for both SSL and FSL cases.   

\begin{table}[]
\centering
{%
\scalebox{0.65}{
\begin{tabular}{clclclcc}
\hline
classification &  & semantic &  & pseudo-label &  & MSCOCO (FSL) & Clipart (SSL) \\ \hline
 &  & $\checkmark$ &  &  &  & 66.41 & 43.39 \\
$\checkmark$ &  &  &  &  &  & 54.55 & 42.74 \\
 &  & $\checkmark$ &  & $\checkmark$ &  & 67.68 & 48.89 \\
$\checkmark$ &  & \multicolumn{1}{l}{} &  & $\checkmark$ &  & 57.25 & 48.57 \\
$\checkmark$ &  & $\checkmark$ &  & $\checkmark$ &  & 66.41 & 48.65 \\ \hline
\end{tabular}%
}}
\caption{\small \putouralg's performance with different loss settings in SSL and FSL regimes.\vspace{-5mm}}
\label{tab:loss_ablations_ssl_and_fsl}
\end{table}

\begin{figure*}[t]
    \centering
    \includegraphics[width=\textwidth]{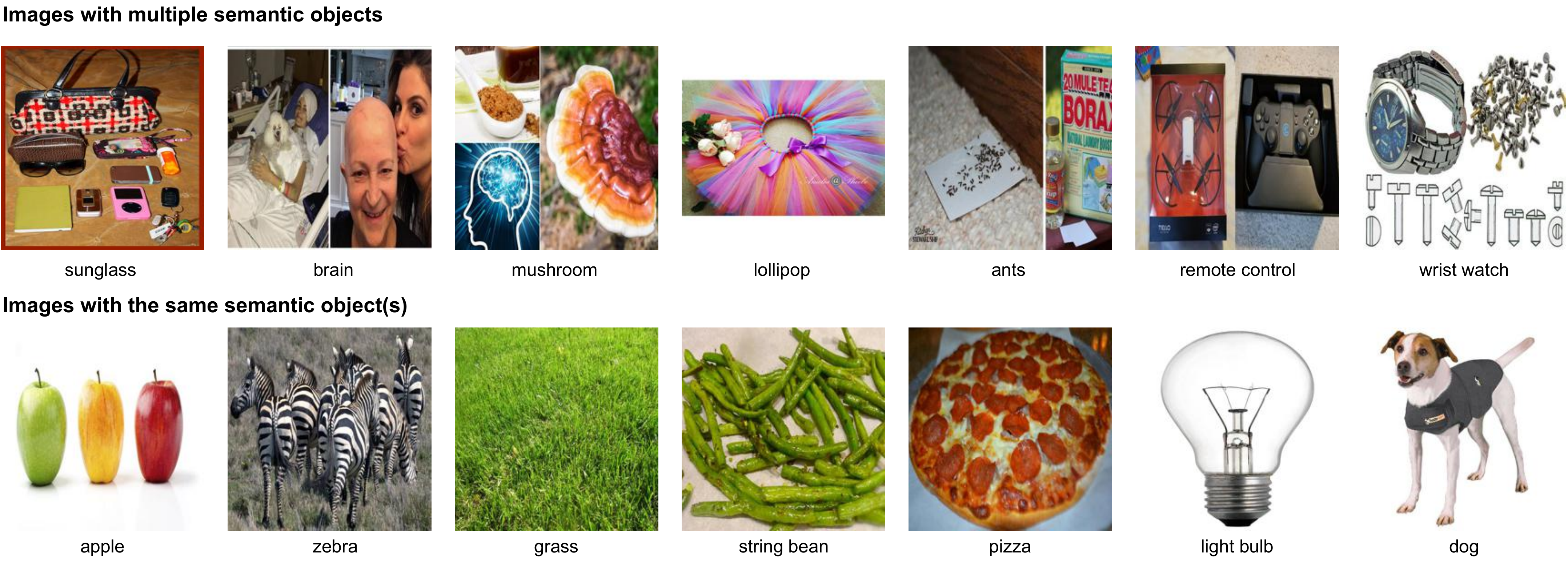}
    \caption{We display a subset of the top (top) and least (bottom) ranked images based on their large crops pseudo-label disagreement rate. We observe that the instances with high disagreement rate are those which contain multiple semantic objects, while instances with the same semantic object tend to suffer less from inconsistencies. \vspace{-4mm}}
    \label{fig:multiple_similar_semantics}
\end{figure*}

\smallskip
\noindent \textbf{Source of semantics. } 
Here, we examine different alternatives to obtain the class label embeddings ($\Omega$). The first choice as in the seminal work of~\cite{frome2013devise} is to simply use the word embeddings of the class labels (such as Word2Vec~\cite{mikolov2013_word2vec} or Glove~\cite{pennington2014glove}) to ground the semantic relations between classes. Word embeddings are learnt in an unsupervised manner to capture contextual relations between words based on their co-occurence statistics in a large corpus of text. This results in vectors which capture contextual similarity but not necessarily visual similarity. Alternatively, Nassar \etal~\cite{nassar2021all} suggested to use knowledge graphs~\cite{speer2016_conceptnet} to adjust word embeddings in a way which also correlates with visual similarity. While the two methods work reasonably well, they both suffer from a coverage issue because they use a predefined set of vocabulary and so it is common that some of the target class labels do not exist in their vocabulary. Accordingly, we suggest in our work to use a pretrained language model using sub-word tokenization such as BERT~\cite{devlin2018bert} and its variants. Such models alleviate out-of-vocabulary problems by operating on parts of words (\ie subwords) instead of words. Hence, to examine these different alternatives, we compare \putouralg's performance when using each of them. We fix the learning task to the MSCOCO FSL task where we run 100 FSL test episodes with different choices of embeddings: 1) Glove word embeddings~\cite{pennington2014glove}, 2) Knowledge graph embeddings~\cite{nassar2021all}, 3) A multilingual BERT-based sentence encoder\footnote{\scriptsize distiluse-base-multilingual-cased-v1 in \\ \url{https://www.sbert.net/docs/pretrained_models.html}}, and 4) a praphrase language model~\cite{song2020mpnet}. We respectively obtain the following top-1 average performance on the 100 FSL MSCOCO episodes (using the same random seed for the episode generation): 63.52\%, 64.12\%, 66.67\%, and 67.68\%. Accordingly, we select the paraphrase model as the default option for \putouralg.

\subsection{Design alternatives for Multi-crop pseudo-labelling}
As discussed in Sec.~\ref{subsec:method_multi_crop_pseudo_label}, we explored different design choices to aggregate the multi-crop losses. We present a comparison between the different design choices in Tab.~\ref{tab:ablations_multi}. Specifically, $\Lcal_{multi}^i$ can be calculated as \textit{pair-wise average soft} pseudo-label loss as in our method (in the main text), or as the \textit{pair-wise average hard} pseudo-label loss, \ie by replacing $(\pv^j_{t})$ with  $\arg\max (\pv^j_{t})$. Alternatively, a \textit{single average soft} pseudo-label can be obtained based on all the teacher crops then used as a soft target against all the student crops predictions, \ie \begin{align}\label{eq:single_soft}
\Lcal_{multi}^i &= \frac{1}{| \mathcal{S}^i|} \sum_{ \Tilde{\uv^j}_s \in \mathcal{S}^i }  
-  \pv^i_{t} \log \pv^j_s, \\ 
\text{where, }\quad \label{eq:aggregate} \pv^i_{t} &= \frac{1}{| \mathcal{T}^i|} \sum_{ \Tilde{\uv^j}_t \in \mathcal{T}^i }  
-  \pv^j_{t}
\end{align}
or a \textit{single average hard} pseudo-label, \ie by replacing $(\pv^i_{t})$ with  $\arg\max (\pv^i_{t})$ in Eqn.~\ref{eq:single_soft}, or finally a \textit{single majority hard} pseudo-label by applying a majority vote on the hard predictions of all the teacher crops, \ie using  $\pv^i_{t} = \text{majority}(\arg \max(\pv^j_{t})) | \Tilde{\uv^j}_t \in \mathcal{T}^i$ instead of Eqn.~\ref{eq:aggregate}.

\begin{table*}[]
\centering
\resizebox{\textwidth}{!}{%
\begin{tabular}{lcc|lccc|lccc}
\toprule[0.15em]
 & \textbf{real 2-shot} & \textbf{clipart 2-shot} & \textbf{} & \multicolumn{1}{l}{\textbf{}} & \textbf{real 2-shot} & \textbf{clipart 2-shot} & \textbf{} & \multicolumn{1}{l}{\textbf{}} & \textbf{real 2-shot} & \textbf{clipart 2-shot} \\
\textbf{Aggregation strategy} & \multicolumn{1}{l}{\textbf{}} & \multicolumn{1}{l|}{\textbf{}} & \textbf{} & \multicolumn{1}{l}{\textbf{Small crops count}} & \multicolumn{1}{l}{\textbf{}} & \multicolumn{1}{l|}{\textbf{}} & \textbf{} & \multicolumn{1}{l}{\textbf{Momentum}} & \multicolumn{1}{l}{} & \multicolumn{1}{l}{} \\ \midrule[0.15em]
pair-wise average soft & 58.79 & 48.65 &  & 0 & 54.26 & 46.97 &  & 0 & 27.84 & 38.44 \\
pair-wise average hard & 57.22 & 46.25 &  & 4 & 56.12 & 47.9 &  & 0.9 & 51.02 & 46.02 \\
single average soft & 55.89 & 45.98 &  & 6 & 58.57 & 48.57 &  & 0.95 & 55.38 & 46.86 \\
single average hard & 54.12 & 45.12 &  & 8 & 57.77 & 48.68 &  & 0.99 & 58.67 & 48.57 \\
single majority hard & 55.95 & 46.43 &  & 10 & 57.97 & 48.42 &  & 0.999 & 57.90 & 30.81 \\ \bottomrule[0.15em]
\end{tabular}%
}
\vspace{3mm}
\caption{\textbf{Further Ablations.} Left: multi-crop loss aggregation strategy. Middle: number of small crops seen by the student model. Right: the effect of teacher momentum ($\gamma$).}
\label{tab:ablations_multi}
\end{table*}

\section{Details of Analysis Experiments and Additional Examples}
In this section we provide more details about the experimental setup for the analysis experiments in Sec.~\ref{sec:analysis}, as well as additional qualitative examples to provide more intuition.

\begin{figure*}[t]
    \centering
    \vspace{-3mm}
    \includegraphics[width=\textwidth]{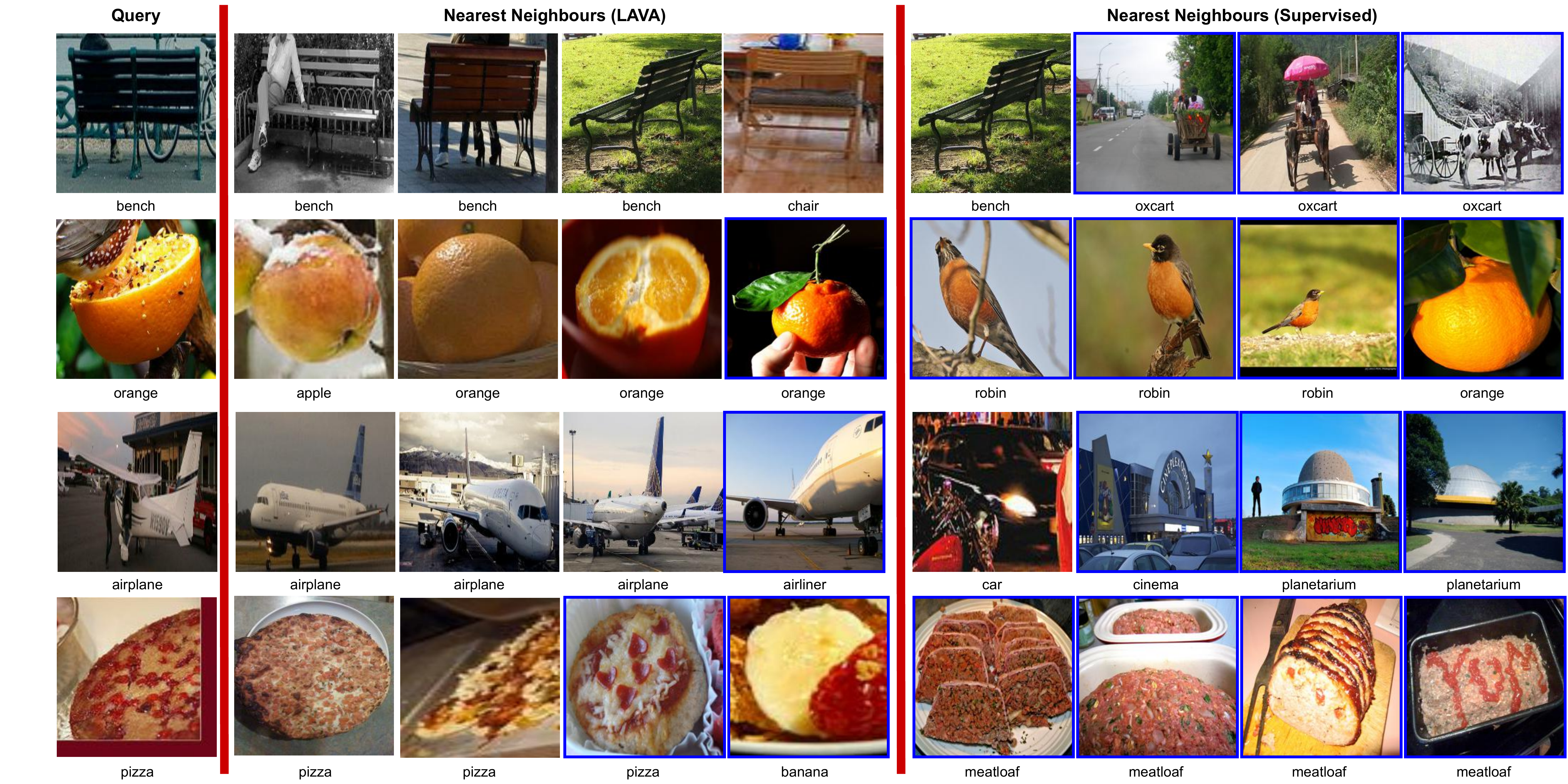}
    \caption{Additional examples of class collapse when transferring from ImageNet to MSCOCO dataset.}
    \label{fig:sup_collapse_mscoco_additional}
\end{figure*}

\subsection{Multi-crop pseudo-labelling analysis} 
Here, we elaborate on the experimental setup for the experiment presented in Sec.~\ref{sec:analysis} - Fig.~\ref{fig:multi_crop_analysis}. We train \putouralg using the \textit{clipart} 2-shot SSL setting and while training, we capture the teacher and student predictions $\pv_t$ and $\pv_s$ for each of the large and small scale crops for every training iteration. Upon convergence (20 epochs of training), we apply argmax on all the captured values to obtain the most dominant ``pseudo-label'' as viewed by the student/teacher based on each of the crops. Subsequently, we use the ground truth labels of the SSL unlabelled instances to calculate the true top-1 accuracy associated with each of the crops and we average it over each epoch. Additionally, we calculate the disagreement rate among the small/large crops as the ratio between the number of unique pseudo-label classes obtained for small/large crops to the total number of small/large crops used. For example using 6 small crops, if the pseudo-labels obtained were (\textit{dog}, \textit{dog}, \textit{dog}, \textit{cat}, \textit{squirrel}, \textit{mouse}) then the disagreement rate is $ = \frac{4}{6} = 0.667$. In Fig.~\ref{fig:multi_crop_analysis}, we report the accuracy based on one of the large crops seen by the student and teacher (denoted as \textit{student} and \textit{teacher} in the legend) together with two small crops seen by the student (denoted as \textit{small1} and \textit{small2}). Even though, we use 6 small crops and 2 large crops during training, we only display the above mentioned subset to avoid clutter. Moreover, we report the disagreement rate among the small crops and the large crops (denoted as \textit{disagree\_small}, \textit{disagree\_large}). 

\noindent Finally, we rank all the training instances based on the disagreement rate among their large crops, averaged over all the training iterations, to examine which images suffer the most and the least from pseudo-labelling inconsistencies due to cropping. We display an additional subset of the top and least ranked images in Fig.~\ref{fig:multiple_similar_semantics}. We observe that the instances with high disagreement rate are those which contain multiple semantic objects; which confirms our intuition about the necessity of the proposed multi-crop pseudo-labelling strategy.

\subsection{Collapse Analysis}
Here we provide further details about the transfer collapse described in Fig.~\ref{fig:sup_collapse}. Doersch \etal~\cite{doersch2020crosstransformers} originally suggested that collapse happens as a result of the supervised pretraining used by most recent FSL methods when training their base learner. Essentially, since the learner is trained to merely classify images into one of a predefined set of classes, the learner encodes information which is useful to discriminate such training classes but discard any other information including that which is useful to generalise to new classes or domains. We further categorise supervision collapse into two types: class and domain collapse as we illustrate below. To visually examine such problem, we follow a protocol inspired by ~\cite{doersch2020crosstransformers}: first, we train a supervised base learner (ViT) on all the instances of the 712 Meta-dataset ImageNet train classes using standard classification cross-entropy loss and following ~\cite{touvron2021training} training procedure. Then, we randomly select 100 instances per each of the 712 train classes together with 1000 query instances from a given target dataset (\eg \textit{MSCOCO} or \textit{clipart}). Subsequently, we use the supervised base learner and \putouralg to obtain the latent representations (\ie $\zv_t$) for each of the sampled images (712 x 100 train + 1000 test query images); and we retrieve the 10 nearest neighbours (based on cosine similarity) for each of the 1000 query images with respect to such representations.  Note that the 1000 images are never seen by either of the models during training and hence we hope that a general visual learner would be able to retrieve, for each query image, neighbours which are at least semantically related. If the learner retrieves a majority of neighbours which belong to one of the train classes for a given query image, it is said that its representations have collapsed to that training class. In Fig.~\ref{fig:sup_collapse} and Fig.~\ref{fig:sup_collapse_mscoco_additional}, we display collapse examples from \textit{MSCOCO} dataset, where we observe that \eg a query ``bench'' instance has collapsed into ``oxcart'' and an ``orange'' instance has collapsed into a ``robin'' in the supervised learner case, while \putouralg retrieves plausible neighbours.  

\begin{figure*}
    \centering
\scalebox{0.85}{
    \includegraphics[width=\textwidth]{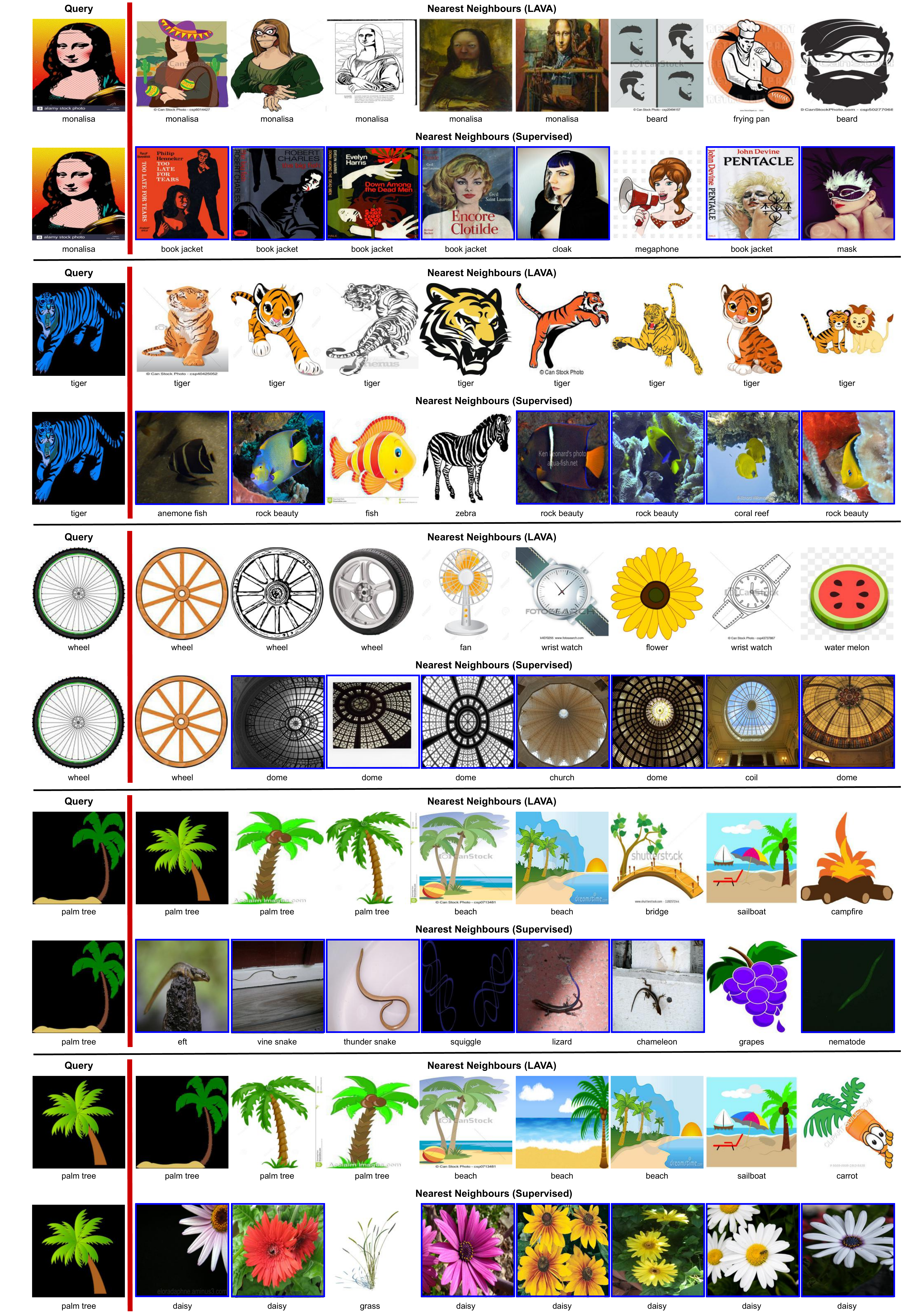}
    }
    \vspace{-1mm}
    \caption{Class collapse examples when transferring from ImageNet to \textit{clipart} dataset. }
    \label{fig:sup_class_collapse}
\end{figure*}

\begin{figure*}
    \centering
    \scalebox{0.85}{
    \includegraphics[width=\textwidth]{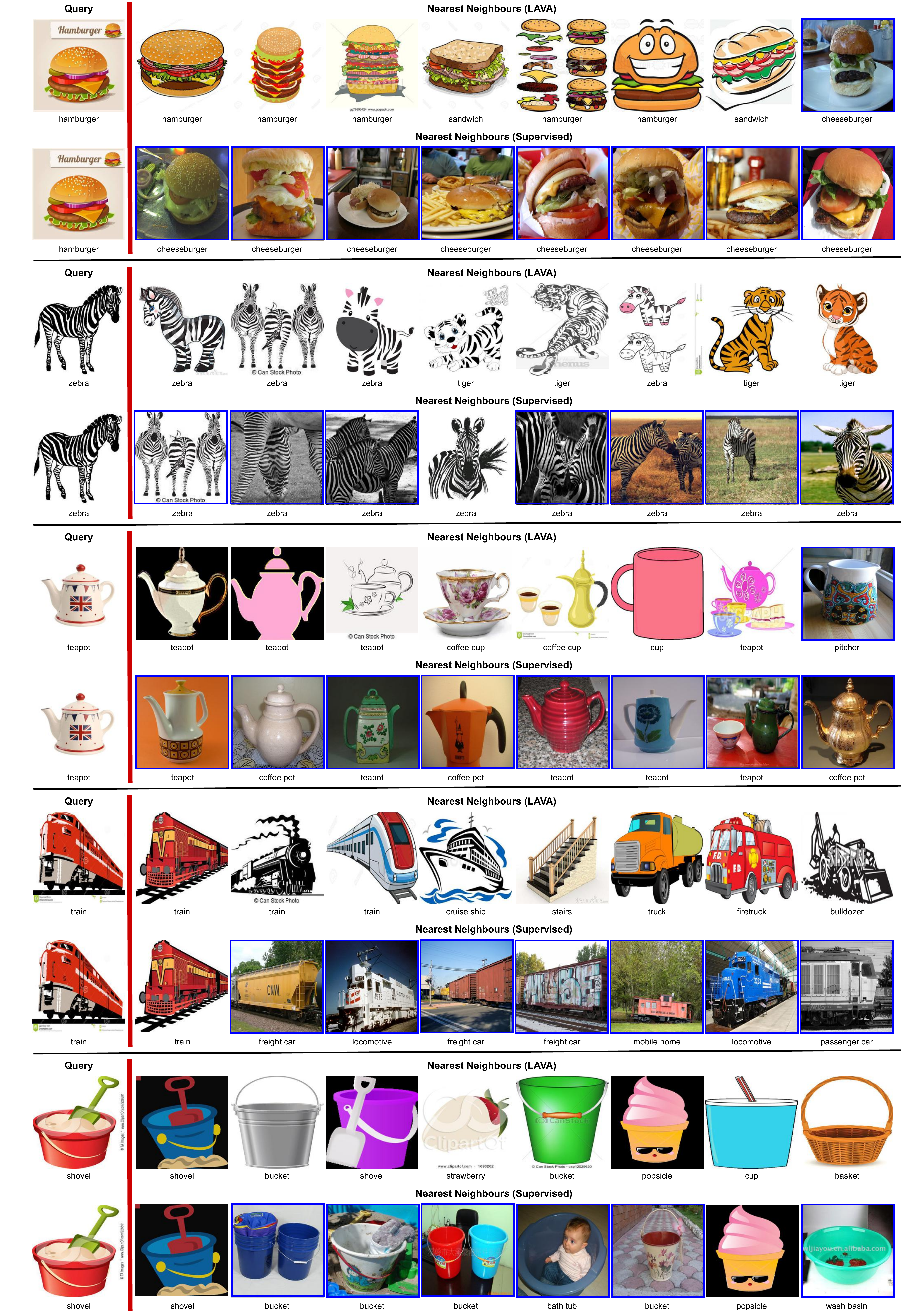}
    }
    \vspace{-1mm}
    \caption{Domain collapse examples when transferring from ImageNet to \textit{clipart} dataset.}
    \label{fig:sup_domain_collapse}
\end{figure*}

To further investigate  collapse in a different visual domain, we conduct a similar experiment but using \textit{clipart} as the target dataset. Interestingly, we witness two types of collapse when the domain also differs, which we denote as ``class collapse'' (Fig.~\ref{fig:sup_class_collapse}) and ``domain collapse'' (Fig.~\ref{fig:sup_domain_collapse}). The former is similar to what was described earlier where the majority of retrieved neighbours belongs to a semantically different train class which shares superficial similarity with the query image. While the latter is when the retrieved neighbours belong to a semantically similar train class albeit in the training visual domain rather than the target domain. To elaborate, examining Fig.~\ref{fig:sup_class_collapse}, we see ``class collapse'' where a ``monalisa'' clipart instance collapses to ``book jacket'' ImageNet class (top section); and a ``wheel'' instance collapses to ``dome'' (third section from top). Whereas in Fig.~\ref{fig:sup_domain_collapse}, we witness a ``domain collapse'' where \eg ``hamburger'' clipart instances collapses to ``cheese burger'' which is semantically related but belongs to the ``real'' visual domain rather than the target ``clipart'' domain.

\clearpage

{\small
\bibliographystyle{ieee_fullname}
\bibliography{egbib}
}

\end{document}